\documentclass[letterpaper, 10 pt, conference]{ieeeconf}  

\IEEEoverridecommandlockouts                              

\usepackage{graphics} 
\usepackage{graphicx}

\usepackage{epsfig} 
\usepackage{amsmath} 
\usepackage{amssymb}  
 \usepackage{algorithm}
\usepackage[compatible]{algpseudocode}
\algnewcommand\AAND{\textbf{ and }}
\algnewcommand\Or{\textbf{ or }}
\usepackage{color}
\usepackage{citesort}
\usepackage{url}

\DeclareMathAlphabet{\pazocal}{OMS}{zplm}{m}{n}

\DeclareMathAlphabet{\mathpzc}{OT1}{pzc}{m}{it}

\usepackage{array}
\newcolumntype{C}[1]{>{\centering\arraybackslash}p{#1}}
\newcolumntype{M}[1]{>{\raggedright\arraybackslash}p{#1}}

\usepackage{array} 
\newcolumntype{L}[1]{>{\raggedright\let\newline\\\arraybackslash\hspace{0pt}}m{#1}}	
\newcolumntype{S}[1]{>{\centering\let\newline\\\arraybackslash\hspace{0pt}}m{#1}}
\newcolumntype{R}[1]{>{\raggedleft\let\newline\\\arraybackslash\hspace{0pt}}m{#1}}

\makeatletter
\renewcommand*{\@opargbegintheorem}[3]{\trivlist
  \item[\hskip \labelsep{\itshape #1\ #2}] \textit{(#3)}\ }
\makeatother

\title{\LARGE \bf
Resource-aware Online Parameter Adaptation for Computationally-constrained Visual-Inertial Navigation Systems
}

\author{Pranay Mathur$^1$, Nikhil Khedekar$^2$, and Kostas Alexis$^2$
\thanks{$^1$~Pranay Mathur is an undergraduate student pursuing Electronics and Instrumentation Engineering at Birla Institute of Technology and Science (BITS) Pilani Goa Campus, India {\tt\small f20170487@goa.bits-pilani.ac.in} }
\thanks{$^2$~The authors are with the Autonomous Robots Lab, Norwegian University of Science and Technology, O. S. Bragstads Plass 2D, 7034, Trondheim, Norway.}
}

\begin{document}

\maketitle
\thispagestyle{empty}
\pagestyle{empty}

%

\begin{abstract}
In this paper, a computational resources-aware parameter adaptation method for visual-inertial navigation systems is proposed with the goal of enabling the improved deployment of such algorithms on computationally constrained systems. Such a capacity can prove critical when employed on ultra-lightweight systems or alongside mission critical computationally expensive processes. To achieve this objective, the algorithm proposes selected changes in the vision front-end and optimization back-end of visual-inertial odometry algorithms, both prior to execution and in real-time based on an online profiling of available resources. The method also utilizes information from the motion dynamics experienced by the system to manipulate parameters online. The general policy is demonstrated on three established algorithms, namely S-MSCKF, VINS-Mono and OKVIS and has been verified experimentally on the EuRoC dataset. The proposed approach achieved comparable performance at a fraction of the original computational cost. 
\end{abstract}

\section{INTRODUCTION}\label{sec:intro}
Robotic systems and especially Micro Aerial Vehicles (MAVs) owe their autonomous capabilities largely due to the progress in the domain of localization and mapping methods capable of running online and onboard. Among the multiple sensor fusion strategies employed, fusing visual cameras, LiDAR sensors and more, the combination of vision and inertial sensors is an appealing solution due to its small size, low weight, low cost and demonstrated ability to estimate the robot pose with accuracy and robustness. Major successes in the domain have been demonstrated through important milestones, including the ability to provide reliable odometry in both outdoor~\cite{sfly} and indoor~\cite{Burri2016euroc} settings, reliability in fast navigation~\cite{VINS-Mono,openvins2020icra,MSCKF-VIO}, resilience against visual degradation~\cite{KTIO_ICRA_2019,khattak2019robust,zhao2020tp} and more. The community has particularly focused on robust Visual-Inertial Odometry (VIO) solutions, while selected works aim to address the overall Simultaneous Localization And Mapping (SLAM) problem~\cite{ORBSLAM3_2020,slamTROcadena}. Despite the progress, however, visual-inertial odometry estimation remains computationally expensive and necessitates significant computational resources onboard the flying robot. This in turn prohibits the potential of integrating such systems onboard ultra lightweight systems or the ability to run alongside other computationally expensive processes (e.g., planning) in computationally-constrained systems. 

In response to the above, this work contributes a general policy to automatically adapt the behavior and computational profile of a selected set of successful VIO frameworks in order for them to maintain accuracy and robust performance, while self-adjusting their functionality to best fit onboard computationally-constrained systems and respond to real-time changes in the overall available resources as other threads run simultaneously. More specifically, the proposed approach on Computational Resources-aware Visual-Inertial Odometry Scheduling (CRVIOS), is demonstrated in connection to three established VIO methods, namely a) Stereo Multi-State Constraint Kalman Filter  (S-MSCKF) VIO~\cite{MSCKF-VIO}, b) Robust and Versatile Monocular Visual-Inertial State Estimator (VINS-Mono)~\cite{VINS-Mono} and c) Open Keyframe-based Visual-Inertial SLAM (OKVIS)~\cite{OKVIS}. 

For all three methods, we define a unifying policy, that given different computing systems and online profiling of available resources, proposes selected changes on the vision front-end and optimization back-ends of these methods both for compile-time variables such as the image resolution, and online real-time manipulation of parameters including the number of features tracked, the number of iterations in the optimization steps and the ratio of images processed. The method further relates its online parameter manipulation to the motion dynamics experienced by the robot at any given time. The proposed VIO scheduling functionality achieves almost identical performance to the original methods but often at a fraction of the computational cost hence enhancing the ability to be deployed onboard extremely lightweight, power-efficient processing boards. 

To demonstrate the potential and capabilities of the method, we present a set of experimental case studies. Exploiting the widely-adopted EuRoC dataset~\cite{Burri2016euroc}, we demonstrate CRVIOS for the selected VIO methods on both a) a low-key x86 architecture processor, and b) a low-power ARM system. Specifically, an Intel Core i3-4010U with two cores at $1.7\textrm{GHz}$ and a Raspberry Pi 4B integrating a Quad core Cortex-A72 (ARM v8) 64-bit SoC at $1.5\textrm{GHz}$ are considered. For both systems, we run tests by launching additional processes in real-time, thus modifying the actual available resources, and evaluating the resilience of CRVIOS in delivering accurate results by adjusting the parameters of the underlying estimation frameworks. 

The remainder of this paper is organized as follows: Section~\ref{sec:related} outlines related work. The proposed approach is detailed in Section~\ref{sec:approach} with the implementation details provided in Section~\ref{sec:implementation}. Evaluation studies are detailed in Section~\ref{sec:experiments}, followed by conclusions drawn in Section~\ref{sec:concl}. 


%
%

\section{RELATED WORK}\label{sec:related}
With respect to how different approaches lead to reduced computational cost, most computationally lightweight VIO methods can be broadly classified into two categories. The first relies heavily upon implementation and targets lowering the complexity of the computations or a reduction in time taken by reorganization, pipelining and parallelization~\cite{Zhang2017VisualInertialOO}. The other relies upon researchers altering a) the front-end by choosing to process only selective information-rich features and landmarks~\cite{davison2002activevision,strasdat2009useful} or b) the back-end by graph sparsification~\cite{graphSparsification,smartFactors}. The authors in~\cite{fontan2020info} propose an entropy-driven metric to select only the most informative measurements and achieve a $10\times$ reduction in computation, although the method has been developed for use in AR/VR glasses and well-lit settings. Certain keyframe-based methods discard all other measurements and only process measurements from a subset of spatially distributed camera poses~\cite{Klein2007ParallelTA,konolige2004frameslam}. 
Adjusting the sensitivity of feature detection while simultaneously altering the back-end to switch between EKF-based SLAM and MSCKF~\cite{MSCKF} to fully utilize the computational resources was shown in~\cite{li2012resource}. A similar resource-aware approach is assigning different portions of the CPU budget for processing after classifying features on their feature track length~\cite{Kottas2014ARV}. A different approach is lowering the computational complexity of individual steps involved in the SLAM process~\cite{Paz2008EKFOn}. 
The work in this paper specifically focuses on creating a policy that requires minimum alteration of the underlying framework of the VIO method and proposes an adaptive strategy to alter the parameters according to an online profiling of the resources available.


\section{PROPOSED APPROACH}\label{sec:approach}
This section details the proposed policy to automatically adapt a set of visual-inertial odometry estimation frameworks such that resilient performance with minimal accuracy degradation is ensured, along with a significant decrease in computational complexity and adaptability to the resources available at any given time during a robot mission. It should be noted that the proposed approach does not exploit factors that are specific to any certain VIO framework, rather a set of ubiquitously applicable methods that could be extended to multiple visual-inertial odometry algorithms are presented. 

\subsection{Selected Visual-Inertial Odometry Baselines}

The selection of baseline VIO algorithms, VINS-Mono~\cite{VINS-Mono}, OKVIS~\cite{OKVIS}, and S-MSCKF~\cite{MSCKF-VIO}, was made based on two criteria, namely a) their performance on established datasets (e.g., EuRoC~\cite{Burri2016euroc}), and b) their diversity and utilization of distinct functional principles such that our approach can be applicable to a diverse set of methods. Firstly, while VINS-Mono and OKVIS utilize a non-linear optimisation approach in their back-end with a sliding window in the former and a set of keyframes in the latter, S-MSCKF demonstrates a filter-based approach. Furthermore, while VINS-Mono tracks features detected in images from a monocular camera, OKVIS and S-MSCKF are both stereo based.
 
\begin{figure}[h!]
\centering
    \includegraphics[width=0.99\columnwidth]{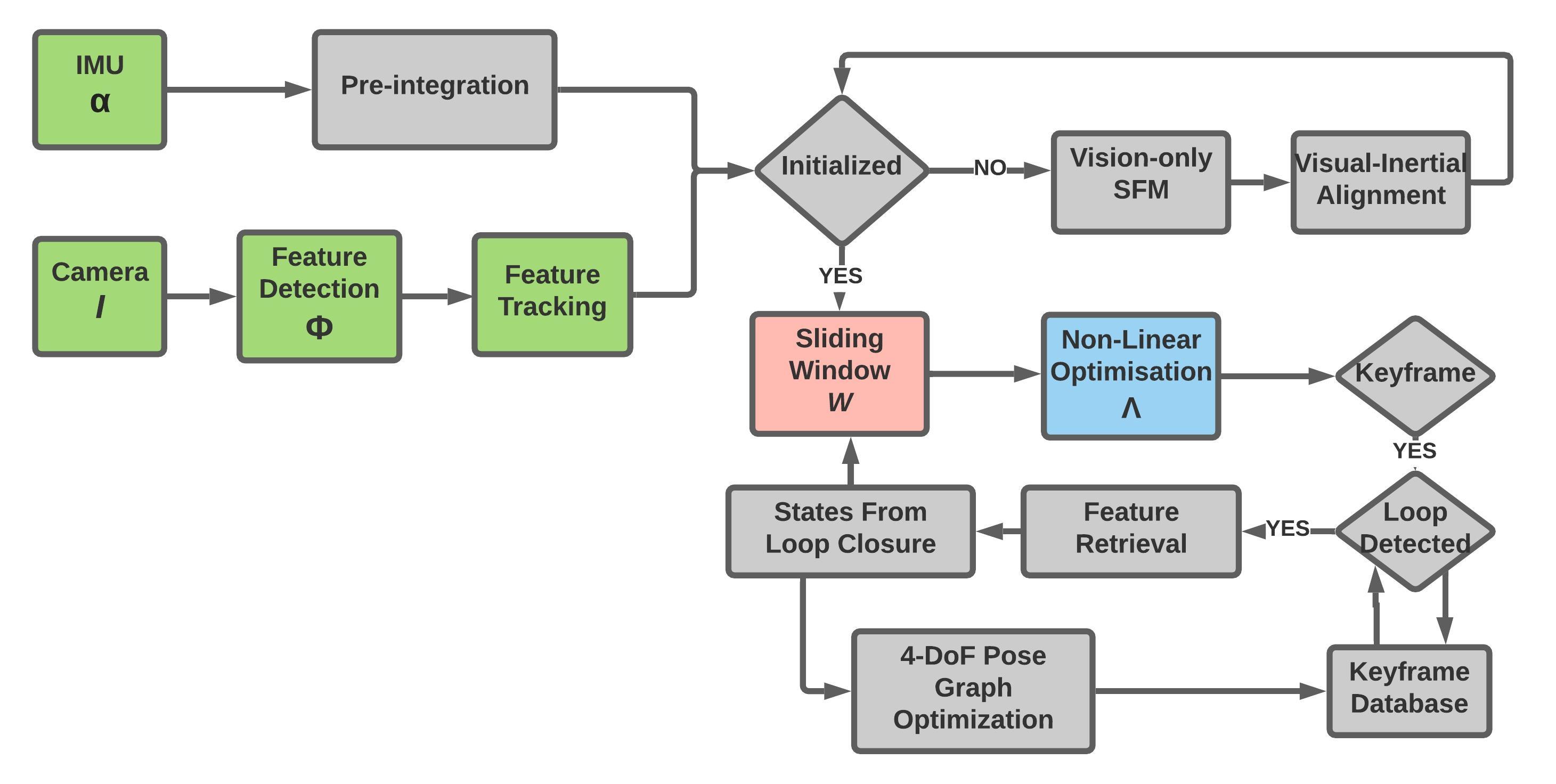}
\caption{VINS-Mono block diagram. Modifiable parameters are indicated by the color of the blocks: Green indicates parameter changes during execution, pink indicates parameter changes at compile time, blue indicates setting the parameters prior to execution and grey indicates no changes.}\label{fig:VINS-block-diagram}
\vspace{-2ex}
\end{figure}

\subsubsection{VINS-Mono}

The ``Robust and Versatile Monocular Visual-Inertial State Estimator'' (VINS-Mono)~\cite{VINS-Mono}, is a monocular visual-inertial 6 DoF state estimator based on tightly-coupled sliding window non-linear optimization. Loosely-coupled sensor fusion is performed to initialize the estimator from an unknown initial state. Subsequently, preintegration~\cite{imu_pre-integration} is performed on IMU measurements before being added to the optimization, and a tightly-coupled formulation for re-localisation is proposed. In the vision processing front-end, for every incoming image frame, robust corner features~\cite{GoodFeatures} are detected and tracked using the KLT sparse optical flow algorithm~\cite{KLT-tracker}. A minimum separation is enforced between them to ensure uniform feature distribution. The method also features modules for tightly integrated loop closure and 4 DoF pose-graph optimization. Its basic processing steps are illustrated in Figure~\ref{fig:VINS-block-diagram}.

\begin{figure}[h!]
\centering
    \includegraphics[width=0.99\columnwidth]{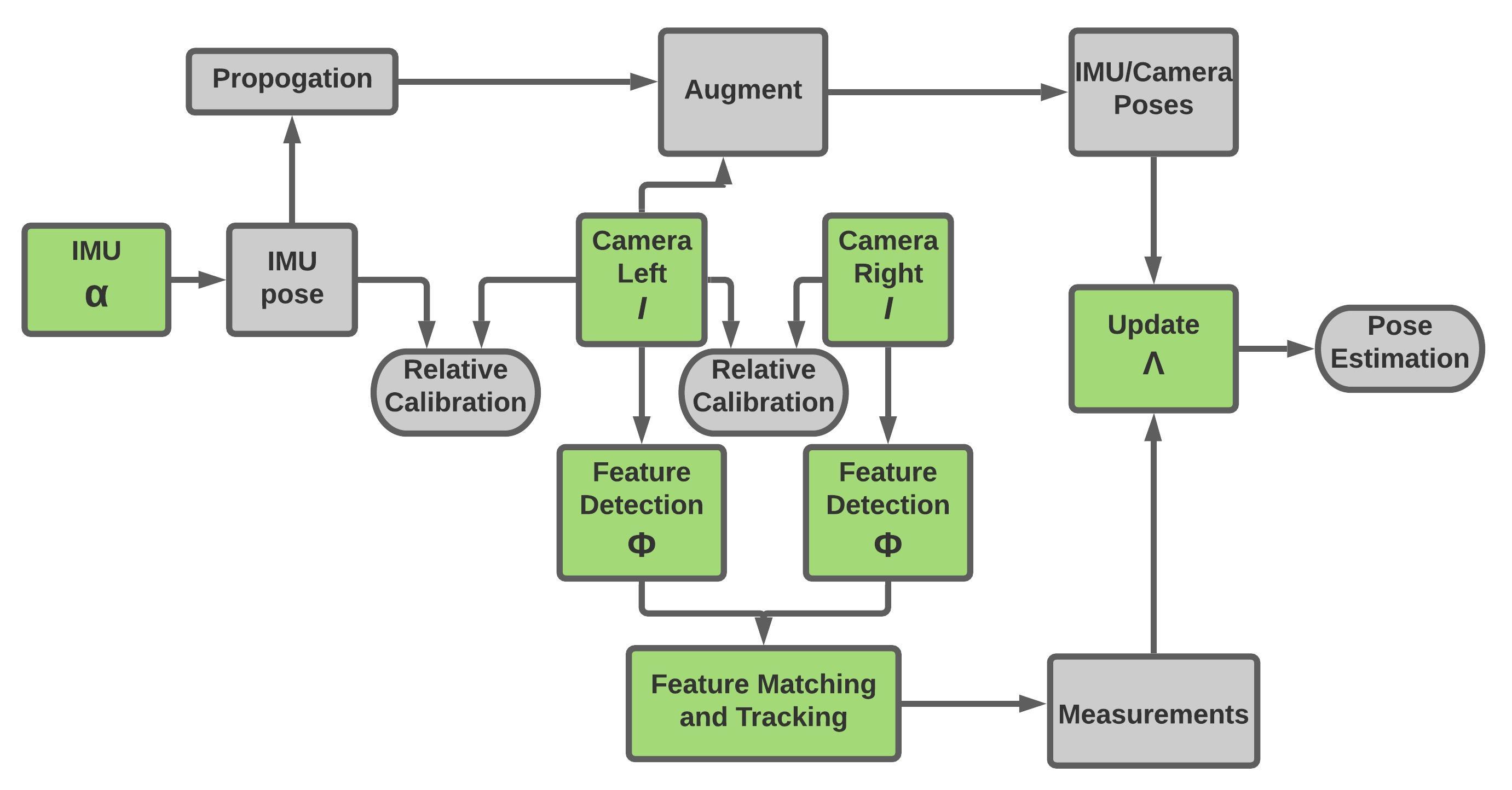}
\caption{S-MSCKF block diagram. Modifiable parameters are indicated by the color of the blocks: Green indicates parameter changes during execution while grey indicates no changes.}\label{fig:S-MSCKF-block-diagram}
\vspace{-2ex}
\end{figure}

\subsubsection{S-MSCKF}

The authors of ``Robust Stereo Visual Inertial Odometry for Fast Autonomous Flight''~\cite{MSCKF-VIO} contributed S-MSCKF, a stereo, filter-based visual-inertial odometry algorithm that uses the established Multi-State Constraint Kalman Filter (MSCKF) originally proposed in~\cite{MSCKF}. The original algorithm~\cite{MSCKF} proposes a measurement model to express the geometric constraints that arise on observing an image feature from multiple camera poses without explicitly adding the features in the state vector. In S-MSCKF, the position of the feature is calculated in the world frame using the least squares method in~\cite{MSCKF} and the estimated camera poses. During the filter update step, two camera states are removed and the feature observations obtained are used for the measurement update. The implementation utilizes the FAST~\cite{FAST} feature detector and tracks features temporally using the KLT optical flow algorithm~\cite{KLT-tracker} which is also used for stereo feature matching. 

\begin{figure}[h!]
\centering
    \includegraphics[width=0.99\columnwidth]{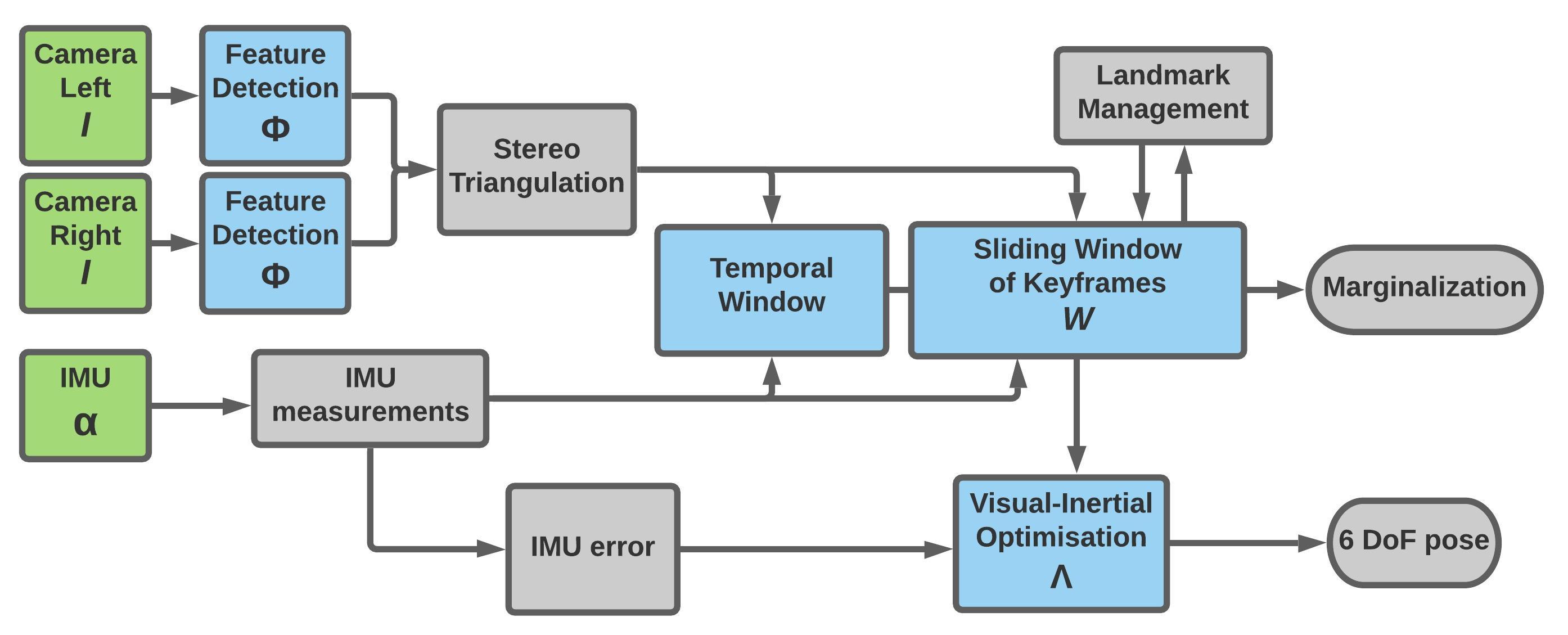}
\caption{OKVIS block diagram. Modifiable parameters are indicated by the color of the blocks: Green indicates parameter changes during execution, blue indicates parameter change prior to execution, and grey indicates no changes.}\label{fig:OKVIS-block-diagram}
\vspace{-2ex}
\end{figure}

\subsubsection{OKVIS}

The ``Open Keyframe-based Visual-Inertial SLAM'' (OKVIS)~\cite{OKVIS} method is based on tightly-coupled fusion of visual information with IMU estimates and uses non-linear optimization over a sliding window of keyframe poses. A probabilistic strategy is used to integrate the IMU error term with the landmark re-projection error resulting in the joint non-linear cost function to be optimized. Partial marginalization of older keyframes in the sliding window is carried-out to maintain a bounded computational complexity. The vision front-end of the algorithm employs a customized Harris corner detector which enforces uniform keypoint distribution and is combined with BRISK descriptor extraction~\cite{BRISK}. For initialization, the last pose propagated by IMU measurements is used to obtain a preliminary uncertain estimate of the state.

\subsection{Tracking of Computational Resources}

The proposed resource-aware online parameter adaptation of VIO strategies, exploits both prior information for the overall CPU capabilities and online monitoring of CPU core usage during the operation of the robotic system. An initial specification of the CPU that details the number of cores, $\mu$, and the maximum and minimum clock speeds, $\nu_{\max}$ and $\nu_{\min}$, is acquired. While both the core usage and process usage for every process associated with the algorithm are recorded, only the core usage, $\chi$, is used for our parameter adaptation. This information is acquired and updated at a rate of $1\textrm{Hz}$ using widely available system monitoring tools. When run on multi-core architectures, the core of execution - within the results presented in this work- is fixed to a specified core for ease of resource monitoring.

\subsection{Algorithm Description}

Our proposed approach involves altering key parameters of both the vision front-end and the optimization back-end to achieve a reduction in computational resource usage. 

\subsubsection{Front-End Adaptation}

The computational reduction in the vision front-end for all methods is achieved by a) choosing to process only a subset, $P$, of the total incoming frames, $I$, and b) by changing the maximum number of features, $\Phi_{I_k}$, detected in the frame at a time $k$, $I_k \in P$. The choice of whether frame $I_k$ $\in$ $P$ and thus if is to be processed is determined by an agility metric, $\alpha$, and the core usage, $\chi$, on which the process is executed. The agility is defined using the average values of linear acceleration, $\alpha^{a}_{avg}$, and angular velocity, $\alpha^{\omega}_{avg}$, over a sliding window of fixed length, $l_{\alpha}$. The degree of agility is determined by a comparison with the respective predefined thresholds $\alpha^{a}_T$ and $\alpha^{\omega}_T$. 
We formalize the above in the following equation, where $\alpha^{\omega}_n$ and $\alpha^{a}_n$ denote the angular velocity and linear acceleration values at time $n$, while $k$ denotes the time that the latest measurement is received:

\small
\begin{equation}
\begin{aligned}
    \alpha^{\omega}_{avg} = (1/l_\alpha) \sum_{n=k-l_\alpha}^{n=k} \alpha^{\omega}_n \\
    \alpha^a_{avg} = (1/l_\alpha) \sum_{n=k-l_\alpha}^{n=k} \alpha^{a}_n
\end{aligned}
\end{equation}
\normalsize

To counter excessive loss of frames in the case of low agility trajectories, a queue over a fixed window of incoming frames is maintained which records whether a frame was processed or dropped and $\alpha^{\omega}_T$ and $\alpha^{a}_T$ are adjusted based on whether the number of processed frames in the queue, $N(P)$, exceeds or falls below fixed thresholds of minimum, $C_0$, and maximum, $C_1$, number of frames to be processed. It is noted that no image frames are dropped during the initialization period. 

Finally, the resolution, $r$, of the images is another factor influencing resource usage and estimation accuracy, and may be altered prior to execution based on the overall computational capabilities at hand. 





\subsubsection{Back-End Adaptation}

The computational reduction and adaptability in the back-end is obtained by altering the maximum number of iterations, $\Lambda$, required for linearization in case of filter-based methods or by the linear solver for non-linear optimisation based methods. In the case of the latter, the size of the sliding window, $W$, is also altered. 

The alterations made can further be classified into three categories depending on when the changes occur: a) prior to compilation b) at initialization and c) during execution (online) and are highlighted in pink, blue and green respectively in Figures~\ref{fig:VINS-block-diagram}-\ref{fig:OKVIS-block-diagram} for the baseline VIO methods. Certain parameters have a major impact in decreasing resource usage when fixed initially, however have a detrimental impact when they are altered online and lead to instability in the system. In the case of OKVIS, all of these parameters are fixed prior to execution. For S-MSCKF both $\Phi$ and $\Lambda$ are altered online. For VINS-Mono, only $\Phi$ is altered online while $W$ and $\Lambda$ are fixed prior to execution. The aforementioned agility-based selective processing of frames is utilized for all methods and thus the system adapts online which frames to process. On the other hand, the image resolution is fixed prior to execution. 

Algorithm~\ref{alg:crvios} outlines the main steps in the proposed approach. Upon initialization, the method provides an initial estimate of all parameters for the VIO method on the basis of the CPU specifications and instantiates a queue, $queue_I$, to record whether the frame $I_k$ is processed or dropped. The choice of altering the maximum number of features detected, $\Phi$, the maximum number of iterations, $\Lambda$, and the extent to which they are altered is determined based on a metric $\Delta = \chi -\chi_T$ corresponding to the difference of current core usage and a fixed threshold.


\begin{algorithm}[H]
    \caption{Resource-Aware Policy Adaptation}
    \label{alg:crvios}
    \begin{algorithmic}[1]
    \renewcommand{\algorithmicrequire}{\textbf{Input:}}
        \REQUIRE $CPU_{spec}$, $\chi_T$, $\alpha_T, \kappa_{0,f},\kappa_{0,p}$
        \STATE $\Phi$, $\Lambda$, $W \leftarrow  \mathbf{ComputeInitialParameters}(CPU_{spec})$
        \STATE $\kappa_f \leftarrow0$ , $\kappa_p\leftarrow0$ , $queue_I\leftarrow 0$
        \FORALL{$I_k \in I$}
            \IF {($\alpha^{\omega}_k<\alpha^{\omega}_T \; and \;\alpha^{a}_k<\alpha^a_T \; and \; \kappa_f>\kappa_{0,f}$)}
                \STATE $queue_I\leftarrow0$ , $\kappa_f\leftarrow0$
                \STATE return
            \ENDIF
            \STATE $\Delta\leftarrow\chi - \chi_T $
            \IF {($\Delta>\Delta_0 \; and \; \kappa_f>\kappa_{0,f}$)}
                \STATE $queue_I\leftarrow0$ , $\kappa_f\leftarrow0$
                \STATE return
            \ELSIF{($\kappa_p>\kappa_{0,p}$)}
                \STATE $\mathbf{UpdateIterations}(\Delta,\Lambda)$
                \STATE $\mathbf{UpdateFeatures}(\Delta,\Phi)$
                \STATE $\kappa_p \leftarrow 0$
            \ENDIF
            \STATE $queue_I\leftarrow 1$
            \WHILE{$(size(queue_I)>=l_I)$}
                \STATE $pop$ $queue_I$
            \ENDWHILE
            \IF{$(sum(queue_I)>C_0)$}
                \STATE Increment $\alpha^{\omega}_T$, $\alpha^{a}_T$
            \ELSIF{$(sum(queue_I<C_1)$}
                \STATE Decrement $\alpha^{\omega}_T$, $\alpha^{a}_T$
            \ENDIF
            \STATE Increment $\kappa_f,\kappa_p$
        \ENDFOR
    \end{algorithmic}
\end{algorithm}

Due to different amounts of reduction in the computational resources being achieved by dropping a frame entirely, reducing the number of iterations or reducing the number of features tracked, these adaptations are made on the basis of different thresholds $\Delta_0$, $\Delta_1$ and $\Delta_2$. Additionally, as dropping several frames could lead to an unstable system, it is also governed by the number of frames processed, $N(P)$, over a fixed window. A count of the number of frames since the last frame drop, $\kappa_f$, is maintained to ensure that if the current frame is dropped, no frame within $\kappa_{0,f}$ subsequent frames may be dropped. To prevent degeneracy due to rapid changes in $\Lambda$ or $\Phi$, no parameter is altered till the number of iterations since the last modification, $\kappa_p$, exceeds a minimum of $\kappa_{0,p}$ after an alteration is made. 

\section{IMPLEMENTATION DETAILS}\label{sec:implementation}
In this section, we review further details of the particular implementation of our strategy. Tuning certain method-specific parameters apart from those detailed as part of our general approach resulted in better performance however these were not changed while evaluating the algorithm on different sequences of the dataset in the interest of fair evaluation. They were fixed to the values which gave the best performance when tested without the resource-aware algorithm. Loop closure was disabled in every case and all VIO methods were compiled with the highest levels of compiler optimization.

For every VIO algorithm, an initial estimate of the parameters is calculated using the maximum clock speed, $\nu_{max}$, and number of cores in the CPU, $\mu$. The domain for this initialization is divided into three regions, namely $R_1$ for single core architectures with $\nu_{max} < 1.2\textrm{GHz}$, $R_2$ for multi-core architectures with $1.2\textrm{GHz} \le \nu_{max} \ge 2.1\textrm{GHz}$ and $R_3$ for multi-core architectures with $\nu_{max}>2.1\textrm{GHz}$. These regions were chosen since the variation of each parameter as a function of the clock speed, $\nu_{max}$ and number of cores in the CPU, $\mu$ followed a similar trend in the given ranges. Upon experimental evaluation an evident change was noted upon moving between regions.



In VINS-Mono, the size of the sliding window, $W$, is varied quadratically in $R_2$, while its value is fixed in $R_1$ and $R_3$. The maximum number of iterations, $\Lambda$, is varied in steps of two across each region. The number of features, $\Phi$, is varied linearly in all regions. 

Since S-MSCKF ensures a uniform distribution of features by ensuring that a certain number of features are detected in every cell of a grid overlayed on the image, the grid dimensions are varied according to the operating region and are fixed prior to execution. The modifications in $\Phi$ are linear in $R_1$ and $R_2$ and fixed in $R_3$, while the modifications in $\Lambda$ are fixed in $R_1$ and linear in $R_2$ and $R_3$. Being a filter-based approach, the size of the sliding window in the backend, $W$ is not applicable.


OKVIS was, by comparison to VINS-Mono and S-MSCKF, computationally expensive and since execution could not be performed on a single fixed core, used two cores instead. Due to OKVIS maintaining a sliding window of IMU-linked temporal frames as well as one of keyframes, two separate window sizes, $W_t$ and $W_{kf}$ were set. It was tested on the x86-based laptop and all parameters were fixed prior to execution with the only online modification being the agility based dropping of frames. 


\newcommand{\comment}[1]{}
\comment{
\begin{table}[]

\caption{\label{tab:ParamTable} Implemented Parameter Values for each method}
\resizebox{\columnwidth}{!}{%
\begin{tabular}{|c|c|c|c|c|c|c|c|c|}
\hline
\multicolumn{9}{|c|}{\textbf{VINS-Mono}} \\ \hline
\textbf{} & \multicolumn{5}{c|}{\textbf{W}} & \multicolumn{2}{c|}{\textbf{$\Phi$}} & \textbf{$\Lambda$} \\ \hline
\multicolumn{1}{|l|}{\textbf{}} & \multicolumn{2}{c|}{\textbf{$p_0$}} & \multicolumn{2}{c|}{\textbf{$p_1$}} & \textbf{$p_2$} & \textbf{$a_0$} & \textbf{$a_1$} & \textbf{$\Lambda_0$} \\ \hline
\multicolumn{1}{|l|}{\textbf{$R_1$}} & \multicolumn{2}{c|}{6} & \multicolumn{2}{c|}{0} & 0 & -40 & 0.1 & 2 \\ \hline
\textbf{$R_2$} & \multicolumn{2}{c|}{16.4} & \multicolumn{2}{c|}{-0.0153} &$5.952*10^{-6}$ & -80 & 0.1 & 4 \\ \hline
\textbf{$R_3$} & \multicolumn{2}{c|}{10} & \multicolumn{2}{c|}{0} & 0 & 170 & 0 & 6 \\ \hline
\multicolumn{9}{|c|}{\textbf{S-MSCKF}} \\ \hline
\textbf{} & \multicolumn{2}{c|}{\textbf{Grid Rows}} & \multicolumn{2}{c|}{\textbf{Grid Columns}} & \multicolumn{2}{c|}{\textbf{$\Phi$}} & \multicolumn{2}{c|}{\textbf{$\Lambda$}} \\ \hline
\textbf{} & \textbf{$p_0$} & \textbf{$p_1$} & \textbf{$p_0$} & \textbf{$p_1$} & \textbf{$a_1$} & \textbf{$a_0$} & \textbf{$b_1$} & \textbf{$b_0$} \\ \hline
\multicolumn{1}{|l|}{\textbf{$R_1$}} & 3 & 0 & 4 & 0 & 0.004 & -0.2 & 0 & 1 \\ \hline
\textbf{$R_2$} & 1.2 & 0.0015 & 2.2 & 0.0015 & 0.0033 & -1 & 0.0045 & -4.3 \\ \hline
\textbf{$R_3$} & 5 & 0 & 6 & 0 & 0 & 5 & 0.018 & -22 \\ \hline
\multicolumn{9}{|c|}{\textbf{OKVIS}} \\ \hline
\textbf{} & \multicolumn{2}{c|}{\textbf{$W_{kf}$}} & \multicolumn{2}{c|}{\textbf{$W_t$}} & \multicolumn{2}{c|}{\textbf{$\Phi$}} & \multicolumn{2}{c|}{\textbf{$\Lambda$}} \\ \hline
 & \multicolumn{2}{c|}{\textbf{$W_{0,kf}$}} & \multicolumn{2}{c|}{\textbf{$W_{0,t}$}} & \textbf{$a_1$} & \textbf{$a_0$} & \textbf{$b_1$} & \textbf{$b_0$} \\ \hline
\multicolumn{1}{|l|}{\textbf{$R_1$}} & \multicolumn{2}{c|}{3} & \multicolumn{2}{c|}{2} & 0 & 70 & 0 & 3 \\ \hline
\textbf{$R_2$} & \multicolumn{2}{c|}{5} & \multicolumn{2}{c|}{3} & 0.078 & -23.33 & 0.0045 & -3.3 \\ \hline
\textbf{$R_3$} & \multicolumn{2}{c|}{7} & \multicolumn{2}{c|}{4} & 0 & 200 & 0.0045 & -4.5 \\ \hline
\end{tabular}
}
\end{table}
}


\section{EVALUATION STUDIES}\label{sec:experiments}
The proposed approach is evaluated on the EuRoC~\cite{Burri2016euroc} dataset in Vicon Room 1 and Vicon Room 2 which includes speeds of upto $0.75\textrm{m/s}$ along with significant motion blur and differences in illumination. The dataset provides stereo WVGA monochrome images at $20\textrm{Hz}$, and temporally synchronized IMU data at $200\textrm{Hz}$ with ground-truth provided by a VICON motion capture system. 

The algorithm is evaluated on the x86 architecture using a dual-core Intel Core i3-4010U CPU with a processor clocked at 1.7 GHz along with a 8GB DDR4 RAM on a laptop, and on the ARM architecture using a quad-core Cortex-A72 (ARM v8) 64-bit SoC at 1.5 GHz processor, with an 8GB LPDDR4 SDRAM on a Raspberry Pi 4B.  
 
The effects of individual parameter variation are calculated by varying a parameter in appropriate step sizes and recording the changes in accuracy and process usage. The best estimate of a parameter is considered to be the value at which the highest accuracy is obtained. Analogously, simultaneous variation is performed to examine evidence of any correlation and combined impact on the accuracy and process usage. 

In the optimization back-end of the VIO algorithm, the parameters altered include the number of iterations, $\Lambda$, and the window size, $W$, both of which yield a significant reduction in process usage. A reduction in the former is accompanied with a negligible decline in accuracy while a steep decline is observed on reduction of the latter. In the vision front-end, the number of features, $\Phi$, has the greatest influence on accuracy while the number of frames processed, $N(P)$, has the maximum influence on process usage.   

For every trial, the resultant trajectory of each VIO method with and without CRVIOS running alongside was recorded and evaluated using the trajectory evaluation toolbox from ~\cite{rpg_traj_eval}. The accuracy is defined by the root mean square error (RMSE) for translation and rotation against ground truth over the entire trajectory. The accuracy and process usage results on the x86-based laptop are shown in Table~\ref{tab:Resultsx86} and on the ARM-based Raspberry Pi 4B in Table~\ref{tab:ResultsARM}. The accuracy with the implementation of our policy is comparable to the original algorithm, however a significant reduction in CPU usage is observed. 
No significant reduction in memory usage was observed.

\begin{figure}[h!]
\centering
    \includegraphics[width=0.99\columnwidth]{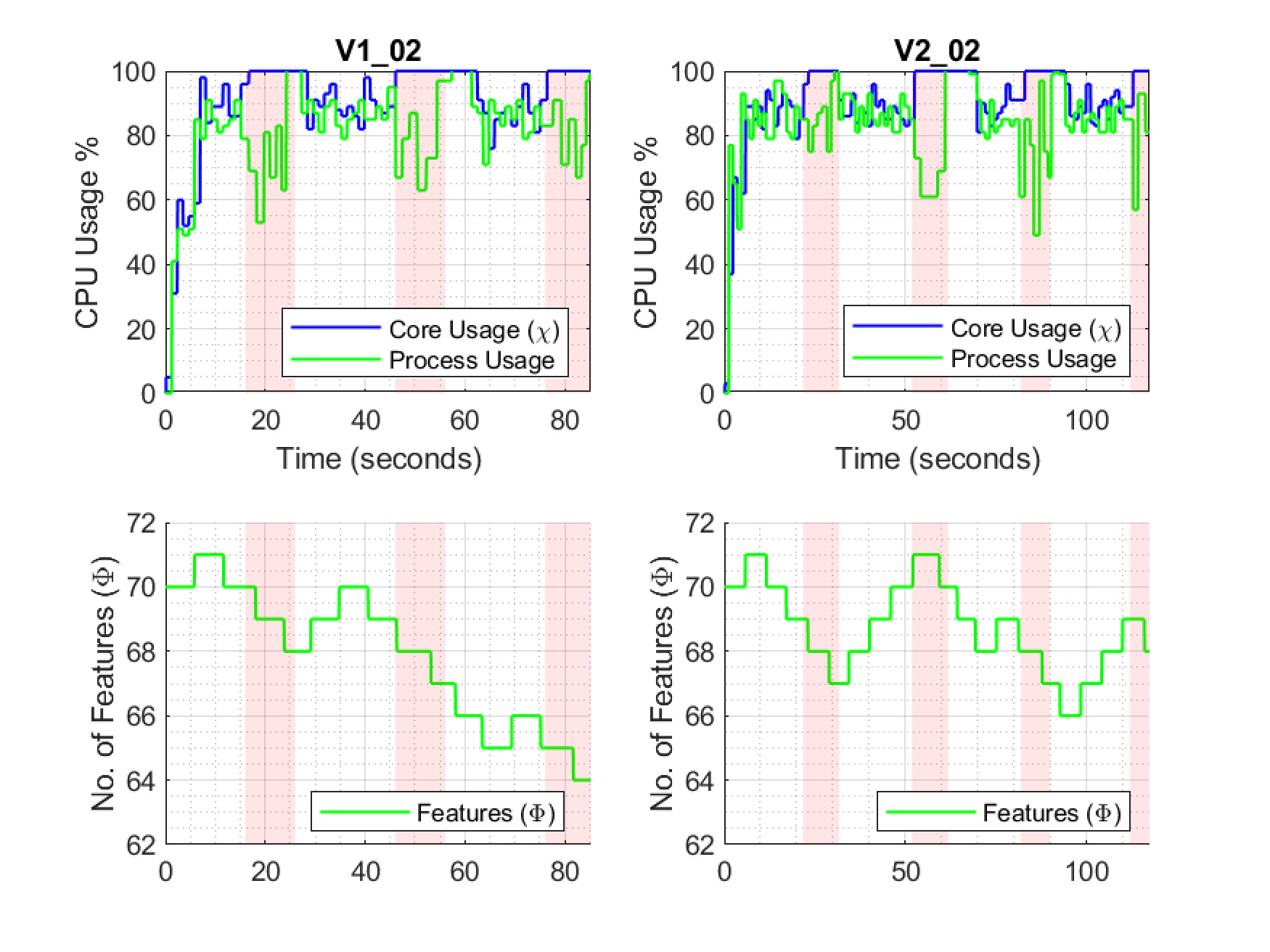}
    \caption{Parameter adaptation on indicative runs of VINS-Mono on the Raspberry Pi 4B on EuRoC sequences V1\_02 (Left) and V2\_02 (Right) with periods of artificial stressing shown in red.}\label{fig:VINS-Mono_graph}
\vspace{-2ex}
\end{figure}

\begin{figure}[h!]
\centering
    \includegraphics[width=0.99\columnwidth]{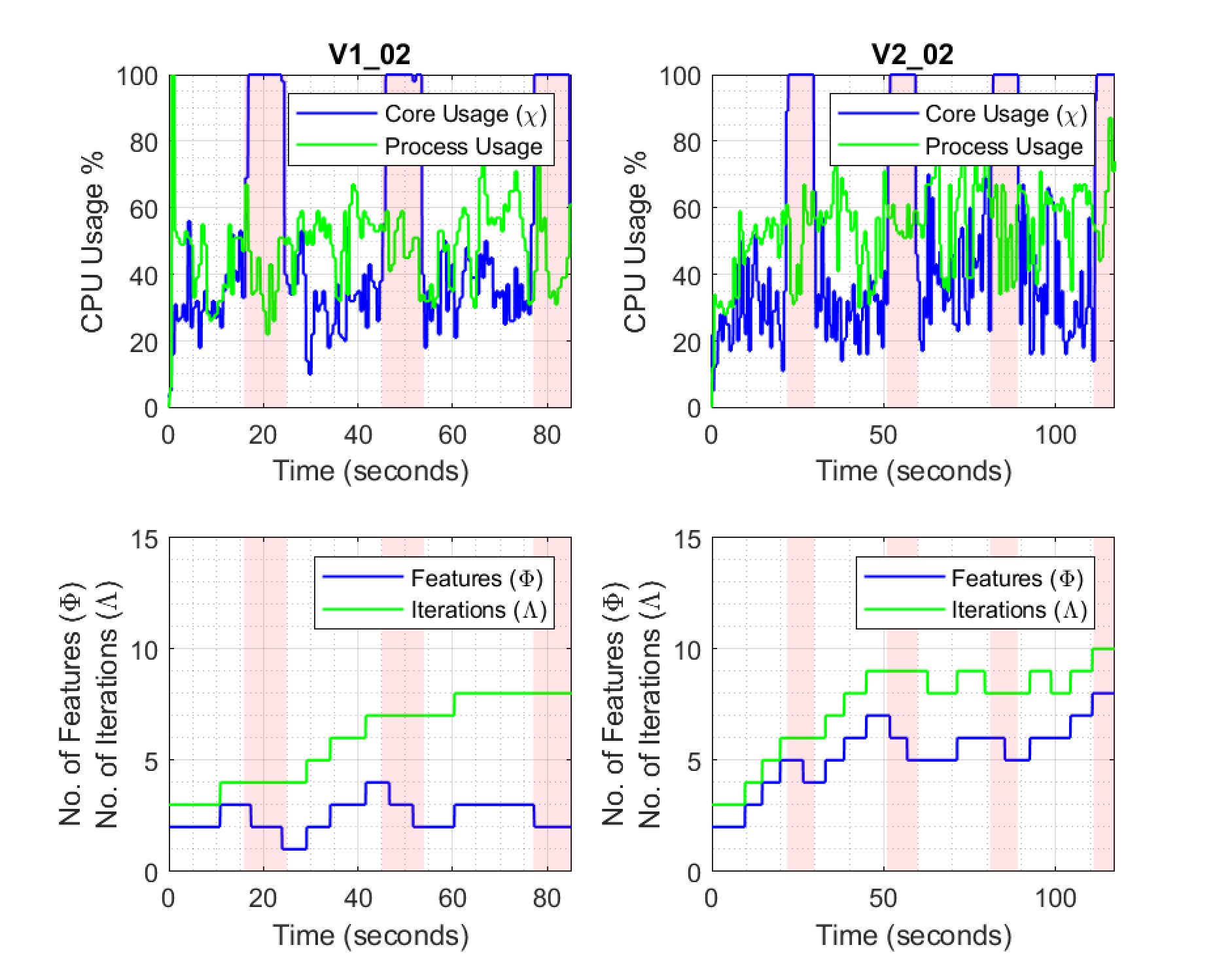}
    \caption{Parameter adaptation on indicative runs of S-MSCKF on the Raspberry Pi 4B on EuRoC sequences V1\_02 (Left) and V2\_02 (Right) with periods of artificial stressing shown in red.}\label{fig:MSCKF_graph}
\vspace{-2ex}
\end{figure}

\begin{figure}[h!]
 \centering
     \includegraphics[width=0.96\columnwidth]{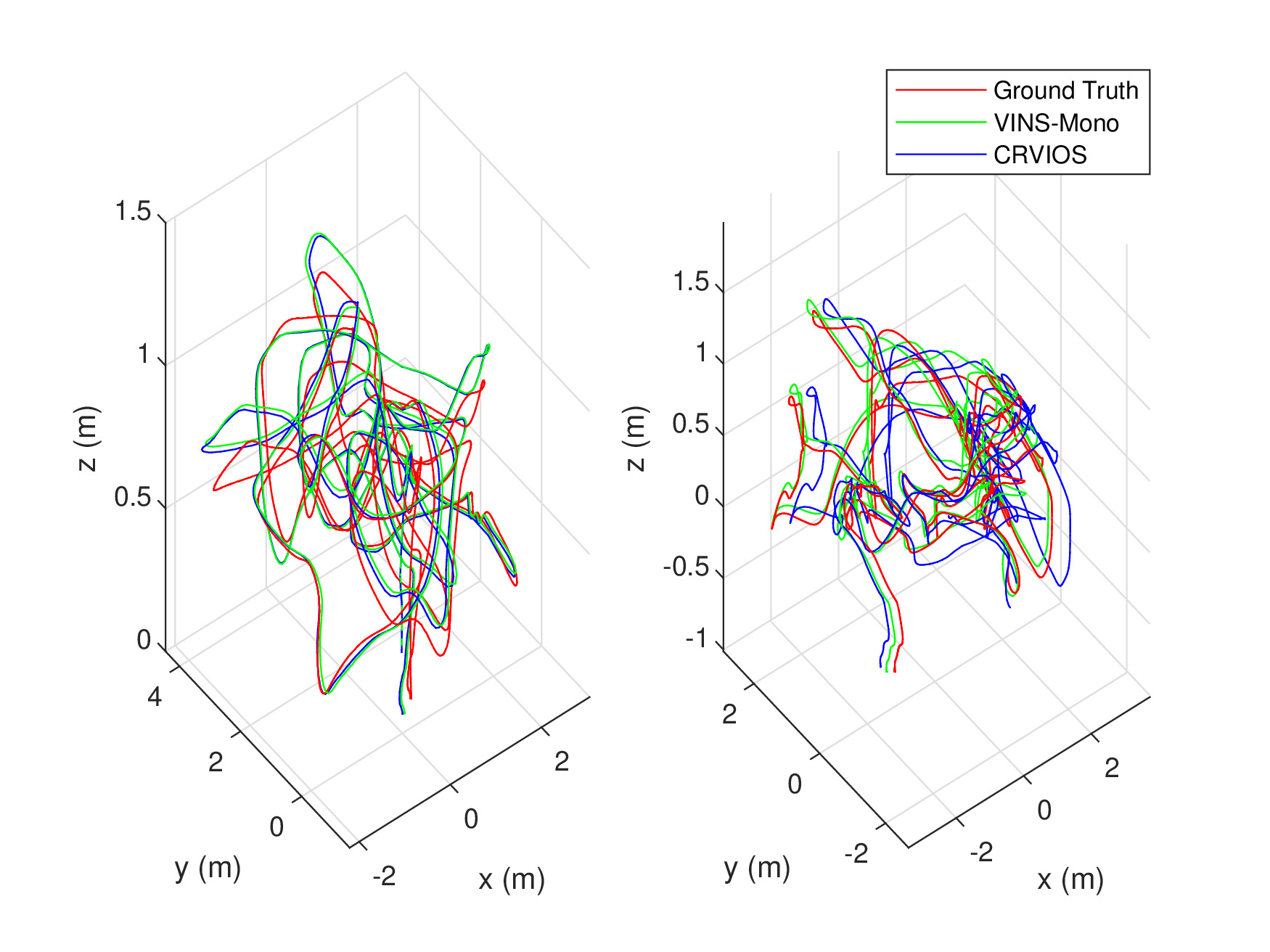}
 \caption{Indicative plots showing the reduced accuracy on running VINS-Mono with CRVIOS on EuRoC sequence V1\_02 (Left) and V2\_02 (Right) on the ARM-based Raspberry Pi 4B. 
 }\label{fig:vins_mono_plots}
 \vspace{-2ex}
 \end{figure}

\begin{figure}[h!]
 \centering
     \includegraphics[width=0.96\columnwidth]{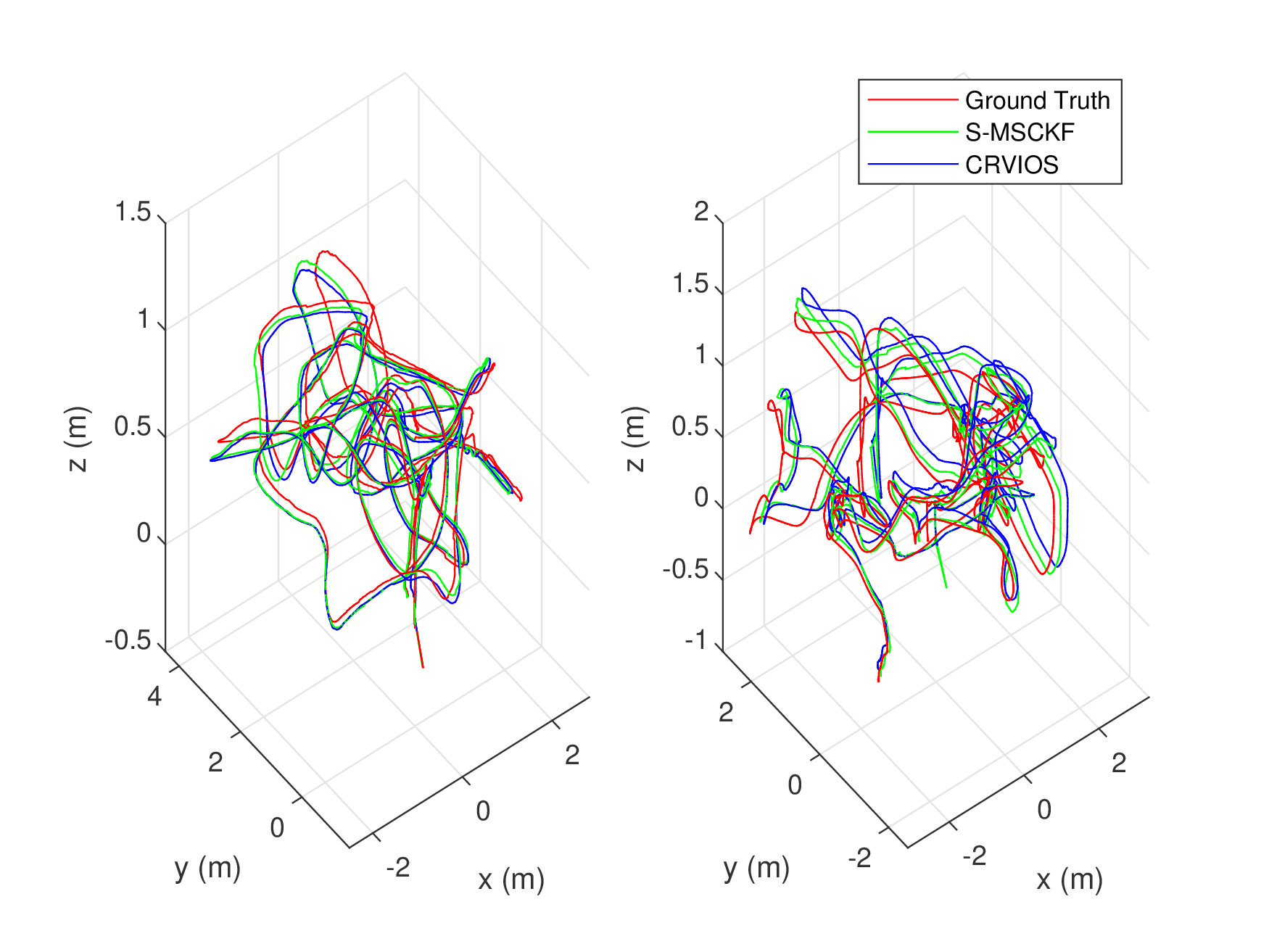}
 \caption{Indicative plots showing the reduced accuracy on running S-MSCKF with CRVIOS on EuRoC sequence V1\_02 (Left) and V2\_02 (Right) on the ARM-based Raspberry Pi 4B. 
 }\label{fig:msckf_plots}
 \vspace{-2ex}
 \end{figure}

To evaluate the reaction of the proposed policy to online changes, all cores of the system are artificially stressed for a fixed duration periodically. Indicative plots for the resulting parameter variation in VINS-Mono and S-MSCKF on the ARM-based Raspberry Pi 4B are provided in Figure~\ref{fig:VINS-Mono_graph} and Figure~\ref{fig:MSCKF_graph} respectively, utilizing the EuRoC sequences V1\_02 and V2\_02. The 3D plots of the corresponding trajectories are provided in Figure~\ref{fig:vins_mono_plots} and Figure~\ref{fig:msckf_plots} respectively.

\begin{figure}[h!]
 \centering
     \includegraphics[width=0.96\columnwidth]{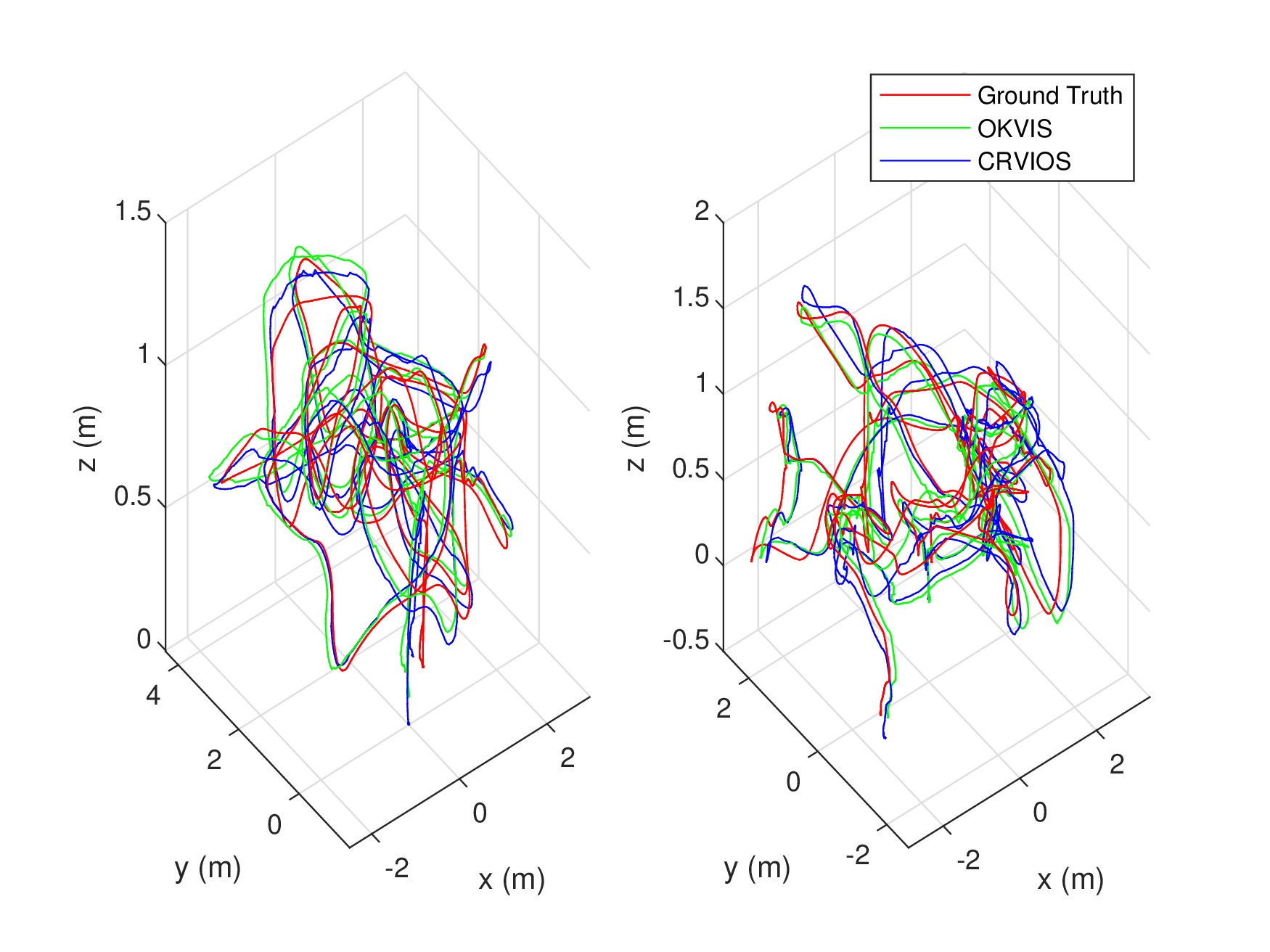}
 \caption{Indicative plots showing the reduced accuracy on running OKVIS with CRVIOS on EuRoC sequence V1\_02 (Left) and V2\_02 (Right) on the x86-based laptop. The corresponding reduction in process usage can be seen in Table~\ref{tab:Resultsx86}.}\label{fig:okvis_plots}
 \vspace{-2ex}
 \end{figure}

Indicative plots comparing OKVIS with and without CRVIOS on EuRoC sequences V1\_02 and V2\_02 are provided in Figure~\ref{fig:okvis_plots} to illustrate the acceptable reduction in accuracy with the reduction in process usage as can be seen in Table~\ref{tab:Resultsx86}. A video recording of a selected subset of the results is available at \url{https://youtu.be/H5gIe418zwc}.

\begin{table*}[ht]
\vspace{3 mm}
\caption{\label{tab:Resultsx86} Accuracy and Process usage changes on x86 - Intel Core i3-4010U @ 1.7GHz}
\centering
\begin{tabular}{|c|c|c|c|c|c|c|c|c|c|}
\hline
\textbf{Dataset} & \textbf{S-MSCKF} & \textbf{CRVIOS} &  & \textbf{VINS-Mono} & \textbf{CRVIOS} &  & \textbf{OKVIS} & \textbf{CRVIOS} &  \\ \hline
 & \multicolumn{2}{c|}{\textbf{Trans.  RMSE (m)}} & \textbf{Change} & \multicolumn{2}{c|}{\textbf{Trans.  RMSE (m)}} & \textbf{Change} & \multicolumn{2}{c|}{\textbf{Trans.  RMSE (m)}} & \textbf{Change} \\ \hline
\textbf{V1\_02} & 0.179 & 0.163 & 0.016 & 0.156 & 0.171 & -0.015 & 0.119 & 0.094 & 0.024 \\ \hline
\textbf{V1\_03} & 0.176 & 0.311 & -0.135 & 0.263 & 0.314 & -0.052 & 0.158 & 0.180 & -0.022 \\ \hline
\textbf{V2\_02} & 0.179 & 0.243 & -0.064 & 0.160 & 0.218 & -0.058 & 0.129 & 0.135 & -0.006 \\ \hline
\textbf{V2\_03} & 0.724 & 0.852 & -0.129 & 0.311 & 0.334 & -0.023 & 0.184 & 0.237 & -0.053 \\ \hline
 & \multicolumn{2}{c|}{\textbf{Rot. RMSE (deg)}} & \textbf{Change} & \multicolumn{2}{c|}{\textbf{Rot. RMSE (deg)}} & \textbf{Change} & \multicolumn{2}{c|}{\textbf{Rot. RMSE (deg)}} & \textbf{Change} \\ \hline
\textbf{V1\_02} & 1.248 & 1.615 & -0.367 & 2.203 & 1.797 & 0.406 & 1.196 & 1.006 & 0.189 \\ \hline
\textbf{V1\_03} & 2.094 & 2.377 & -0.283 & 3.042 & 3.431 & -0.389 & 2.924 & 3.103 & -0.179 \\ \hline
\textbf{V2\_02} & 1.232 & 2.009 & -0.776 & 2.325 & 2.557 & -0.232 & 1.021 & 1.181 & -0.160 \\ \hline
\textbf{V2\_03} & 3.441 & 4.279 & -0.838 & 2.537 & 2.805 & -0.268 & 1.734 & 1.514 & 0.220 \\ \hline
 & \multicolumn{2}{c|}{\textbf{CPU usage \%}} & \textbf{Change} & \multicolumn{2}{c|}{\textbf{CPU usage \%}} & \textbf{Change} & \multicolumn{2}{c|}{\textbf{CPU usage \%}} & \textbf{Change} \\ \hline
\textbf{V1\_02} & 69.7 & 40.8 & 28.9 & 107.6 & 72.2 & 35.4 & 257.0 & 209.0 & 48.0 \\ \hline
\textbf{V1\_03} & 64.8 & 38.6 & 26.2 & 96.0 & 66.9 & 29.1 & 247.0 & 206.0 & 41.0 \\ \hline
\textbf{V2\_02} & 73.6 & 42.2 & 31.4 & 128.6 & 77.6 & 51.0 & 252.0 & 223.0 & 29.0 \\ \hline
\textbf{V2\_03} & 74.5 & 40.6 & 33.9 & 100.9 & 85.0 & 15.9 & 241.0 & 190.0 & 51.0 \\ \hline
\end{tabular}
\end{table*}

\comment{
\begin{table*}[ht]
\vspace{3 mm}
\caption{\label{tab:Resultsx86} Accuracy and Process usage changes on x86 - Intel Core i3-4010U @ 1.7GHz}
\centering
\begin{tabular}{|c|c|c|c|c|c|c|c|c|c|}
\hline
\textbf{Dataset} & \textbf{S-MSCKF} & \textbf{CRVIOS} &  & \textbf{VINS-Mono} & \textbf{CRVIOS} &  & \textbf{OKVIS} & \textbf{CRVIOS} &  \\ \hline
 & \multicolumn{2}{c|}{\textbf{Trans.  RMSE (m)}} & \textbf{\% Change} & \multicolumn{2}{c|}{\textbf{Trans.  RMSE (m)}} & \textbf{\% Change} & \multicolumn{2}{c|}{\textbf{Trans.  RMSE (m)}} & \textbf{\% Change} \\ \hline
\textbf{V1\_02} & 0.179 & 0.163 & 8.914 & 0.156 & 0.171 & -9.875 & 0.119 & 0.094 & 20.304 \\ \hline
\textbf{V1\_03} & 0.176 & 0.311 & -76.896 & 0.263 & 0.314 & -19.702 & 0.158 & 0.180 & -14.082 \\ \hline
\textbf{V2\_02} & 0.179 & 0.243 & -36.021 & 0.160 & 0.218 & -36.214 & 0.129 & 0.135 & -4.553 \\ \hline
\textbf{V2\_03} & 0.724 & 0.852 & -17.760 & 0.311 & 0.334 & -7.253 & 0.184 & 0.237 & -29.036 \\
\hline
 & \multicolumn{2}{c|}{\textbf{Rot. RMSE (deg)}} & \textbf{} & \multicolumn{2}{c|}{\textbf{Rot. RMSE (deg)}} &  & \multicolumn{2}{c|}{\textbf{Rot. RMSE (deg)}} &  \\ \hline
\textbf{V1\_02} & 1.248 & 1.615 & -29.397 & 2.203 & 1.797 & 18.435 & 1.196 & 1.006 & 15.834 \\ \hline
\textbf{V1\_03} & 2.094 & 2.377 & -13.523 & 3.042 & 3.431 & -12.778 & 2.924 & 3.103 & -6.105 \\ \hline
\textbf{V2\_02} & 1.232 & 2.009 & -62.990 & 2.325 & 2.557 & -9.984 & 1.021 & 1.181 & -15.713 \\ \hline
\textbf{V2\_03} & 3.441 & 4.279 & -24.360 & 2.537 & 2.805 & -10.568 & 1.734 & 1.514 & 12.678 \\ \hline
 & \multicolumn{2}{c|}{\textbf{CPU usage \%}} & \textbf{Change} & \multicolumn{2}{c|}{\textbf{CPU usage \%}} & \textbf{Change} & \multicolumn{2}{c|}{\textbf{CPU usage \%}} & \textbf{Change} \\ \hline
\textbf{V1\_02} & 69.700 & 40.800 & 28.900 & 107.600 & 72.200 & 35.400 & 257.000 & 209.000 & 48.000 \\ \hline
\textbf{V1\_03} & 64.800 & 38.600 & 26.200 & 96.000 & 66.900 & 29.100 & 247.000 & 206.000 & 41.000 \\ \hline
\textbf{V2\_02} & 73.600 & 42.200 & 31.400 & 128.600 & 77.600 & 51.000 & 252.000 & 223.000 & 29.000 \\ \hline
\textbf{V2\_03} & 74.500 & 40.600 & 33.900 & 100.900 & 85.000 & 15.900 & 241.000 & 190.000 & 51.000 \\ \hline
\end{tabular}
\end{table*}

\begin{table}[ht]
\centering
\caption{Accuracy and Process usage changes on ARM \\- Cortex-A72 (ARM v8) @ 1.5GHz}
\label{tab:ResultsARM}
\begin{tabular}{|l|r|r|r|}
\hline
\textbf{Dataset} & \multicolumn{1}{l|}{\textbf{S-MSCKF}}      & \multicolumn{1}{l|}{\textbf{CRVIOS}} & \multicolumn{1}{l|}{}                   \\ \hline
                 & \multicolumn{2}{l|}{\textbf{Trans.  RMSE (m)}}                                  & \multicolumn{1}{l|}{\textbf{\% Change}} \\ \hline
\textbf{V1\_02}  & 0.162                                    & 0.148                                & 8.815                                   \\ \hline
\textbf{V1\_03}  & 0.237                                    & 0.223                                & 5.989                                   \\ \hline
\textbf{V2\_02}  & 0.184                                    & 0.183                                & 0.318                                   \\ \hline
\textbf{V2\_03}  & 5.274                                    & 1.726                                & 67.283                                  \\ \hline
                 & \multicolumn{2}{l|}{\textbf{Rot. RMSE (deg)}}                                   & \multicolumn{1}{l|}{}                   \\ \hline
\textbf{V1\_02}  & 1.464                                    & 1.418                                & 3.181                                   \\ \hline
\textbf{V1\_03}  & 2.274                                    & 3.155                                & -38.744                                 \\ \hline
\textbf{V2\_02}  & 1.507                                    & 1.472                                & 2.327                                   \\ \hline
\textbf{V2\_03}  & 7.444                                    & 3.102                                & 58.330                                  \\ \hline
                 & \multicolumn{2}{l|}{\textbf{CPU usage \%}}                                         & \multicolumn{1}{l|}{\textbf{Change}}    \\ \hline
\textbf{V1\_02}  & 83.9                                     & 63.6                                 & 20.3                                    \\ \hline
\textbf{V1\_03}  & 84.2                                     & 70.4                                 & 13.8                                    \\ \hline
\textbf{V2\_02}  & 97.2                                     & 66.9                                 & 30.3                                    \\ \hline
\textbf{V2\_03}  & 71.5                                     & 57.2                                 & 14.3                                    \\ \hline

\textbf{Dataset} & \multicolumn{1}{l|}{\textbf{VINS-Mono}} & \multicolumn{1}{l|}{\textbf{CRVIOS}} & \multicolumn{1}{l|}{}                   \\ \hline
                 & \multicolumn{2}{l|}{\textbf{Trans.  RMSE (m)}}                                  & \multicolumn{1}{l|}{\textbf{\% Change}} \\ \hline
\textbf{V1\_02}  & 0.215                                    & 0.185                                & 14.046                                  \\ \hline
\textbf{V1\_03}  & 0.257                                    & 0.305                                & -18.419                                 \\ \hline
\textbf{V2\_02}  & 0.163                                    & 0.243                                & -49.279                                 \\ \hline
\textbf{V2\_03}  & 0.311                                    & 0.362                                & -16.391                                 \\ \hline
                 & \multicolumn{2}{l|}{\textbf{Rot. RMSE (deg)}}                                   & \multicolumn{1}{l|}{}                   \\ \hline
\textbf{V1\_02}  & 2.257                                    & 2.235                                & 0.972                                   \\ \hline
\textbf{V1\_03}  & 2.983                                    & 3.284                                & -10.099                                 \\ \hline
\textbf{V2\_02}  & 2.615                                    & 2.846                                & -8.827                                  \\ \hline
\textbf{V2\_03}  & 2.537                                    & 3.137                                & -23.645                                 \\ \hline
                 & \multicolumn{2}{l|}{\textbf{CPU usage \%}}                                         & \multicolumn{1}{l|}{\textbf{Change}}    \\ \hline
\textbf{V1\_02}  & 116.3                                    & 100                                  & 16.3                                    \\ \hline
\textbf{V1\_03}  & 131.3                                    & 90.2                                 & 41.1                                    \\ \hline
\textbf{V2\_02}  & 156.6                                    & 100.6                                & 56                                      \\ \hline
\textbf{V2\_03}  & 156.3                                    & 100.6                                & 55.7                                    \\ \hline
\end{tabular}
\end{table}
}

\begin{table}[ht]
\centering
\caption{\label{tab:ResultsARM} Accuracy and Process usage changes on ARM \\- Cortex-A72 (ARM v8) @ 1.5GHz}
\begin{tabular}{|c|c|c|c|}
\hline
\textbf{Dataset} & \textbf{S-MSCKF}          & \textbf{CRVIOS} & \textbf{}       \\ \hline
\textbf{}        & \textbf{Trans.  RMSE (m)} & \textbf{}       & \textbf{Change} \\ \hline
\textbf{V1\_02}  & 0.162                     & 0.148           & 0.014           \\ \hline
\textbf{V1\_03}  & 0.237                     & 0.223           & 0.014           \\ \hline
\textbf{V2\_02}  & 0.184                     & 0.183           & 0.001           \\ \hline
\textbf{V2\_03}  & 5.274                     & 1.726           & 3.549           \\ \hline
\textbf{}        & \textbf{Rot. RMSE (deg)}  & \textbf{}       & \textbf{Change} \\ \hline
\textbf{V1\_02}  & 1.464                     & 1.418           & 0.047           \\ \hline
\textbf{V1\_03}  & 2.274                     & 3.155           & -0.881          \\ \hline
\textbf{V2\_02}  & 1.507                     & 1.472           & 0.035           \\ \hline
\textbf{V2\_03}  & 7.444                     & 3.102           & 4.342           \\ \hline
\textbf{}        & \textbf{CPU usage \%}     & \textbf{}       & \textbf{Change} \\ \hline
\textbf{V1\_02}  & 83.9                      & 63.6            & 20.3            \\ \hline
\textbf{V1\_03}  & 84.2                      & 70.4            & 13.8            \\ \hline
\textbf{V2\_02}  & 97.2                      & 66.9            & 30.3            \\ \hline
\textbf{V2\_03}  & 71.5                      & 57.2            & 14.3            \\ \hline
\textbf{Dataset} & \textbf{VINS-Mono}        & \textbf{CRVIOS} & \textbf{}       \\ \hline
\textbf{}        & \textbf{Trans.  RMSE (m)} & \textbf{}       & \textbf{Change} \\ \hline
\textbf{V1\_02}  & 0.215                     & 0.185           & 0.030           \\ \hline
\textbf{V1\_03}  & 0.257                     & 0.305           & -0.047          \\ \hline
\textbf{V2\_02}  & 0.163                     & 0.243           & -0.080          \\ \hline
\textbf{V2\_03}  & 0.311                     & 0.362           & -0.051          \\ \hline
\textbf{}        & \textbf{Rot. RMSE (deg)}  & \textbf{}       & \textbf{Change} \\ \hline
\textbf{V1\_02}  & 2.257                     & 2.235           & 0.022           \\ \hline
\textbf{V1\_03}  & 2.983                     & 3.284           & -0.301          \\ \hline
\textbf{V2\_02}  & 2.615                     & 2.846           & -0.231          \\ \hline
\textbf{V2\_03}  & 2.537                     & 3.137           & -0.600          \\ \hline
\textbf{}        & \textbf{CPU usage \%}     & \textbf{}       & \textbf{Change} \\ \hline
\textbf{V1\_02}  & 116.3                     & 100             & 16.3            \\ \hline
\textbf{V1\_03}  & 131.3                     & 90.2            & 41.1            \\ \hline
\textbf{V2\_02}  & 156.6                     & 100.6           & 56.0            \\ \hline
\textbf{V2\_03}  & 156.3                     & 100.6           & 55.7            \\ \hline
\end{tabular}
\end{table}

\section{CONCLUSIONS}\label{sec:concl}
In this work, we defined a unifying policy that, given different computing systems, provided optimal initial estimates of parameters and varied them according to the online profiling of available computational resources. Our results show that the proposed VIO scheduling functionality achieves almost identical performance to the original methods but with a major reduction in the computational cost. As demonstrated on two computationally constrained systems, this improves the ability for their deployment. Although we present the effectiveness of our approach on three state-of-the-art VIO algorithms, it can be easily extended to various VIO systems as it targets common functional blocks in both the vision front-end and optimization back-end. Even though there exists an unavoidable trade-off between the computational cost of a VIO system and its accuracy, our contribution endeavours to ensure that this reduction in performance is minimal.  

\bibliographystyle{IEEEtran}
\bibliography{./BIB/CCVIO_ICRA_2021}

\end{document}